\documentclass[10pt,twocolumn,letterpaper]{article}

\usepackage{cvpr}              

\usepackage{graphicx}
\usepackage{amsmath}
\usepackage{amssymb}

\usepackage{kotex} 
\usepackage{xcolor} 
\usepackage{caption} 
\usepackage{multirow, makecell} 
\usepackage{booktabs} 
\usepackage{enumitem}

\usepackage{multirow}
\usepackage{array}
\usepackage{float}
\usepackage{bbm}

  {\begin{list}{}%
          {\setlength{\leftmargin}{#1}}%
          \item[]%
  }
  {\end{list}}

\usepackage{pifont}
\usepackage{pdfrender}
\usepackage{colortbl}
  
\definecolor{color_1}{RGB}{255,0,128}

\newcommand{\ourmodel}{3D-GIF\xspace}

\definecolor{applegreen}{rgb}{0.55, 0.71, 0.0}

\usepackage[pagebackref,breaklinks,colorlinks]{hyperref}

\usepackage[capitalize]{cleveref}
\crefname{section}{Sec.}{Secs.}
\Crefname{section}{Section}{Sections}
\Crefname{table}{Table}{Tables}
\crefname{table}{Tab.}{Tabs.}

\def\cvprPaperID{9943} 
\def\confName{CVPR}
\def\confYear{2022}

\begin{document}


\title{3D-GIF: 3D-Controllable Object Generation via \\Implicit Factorized Representations}

\author{
Minsoo Lee \quad \quad Chaeyeon Chung \quad \quad Hojun Cho \quad \quad Minjung Kim \quad \quad \\ Sanghun Jung \quad \quad Jaegul Choo \quad \quad Minhyuk Sung\\
KAIST, Daejeon, South Korea\\
{\tt\small \{alstn2022, cy\_chung, hojun.cho, emjay73, shjung13, jchoo, mhsung\}@kaist.ac.kr}
}


\twocolumn[{
\renewcommand\twocolumn[1][]{#1}
\maketitle
\begin{center}
    \centering
    \vspace*{-0.7cm}
    \includegraphics[width=1\linewidth]{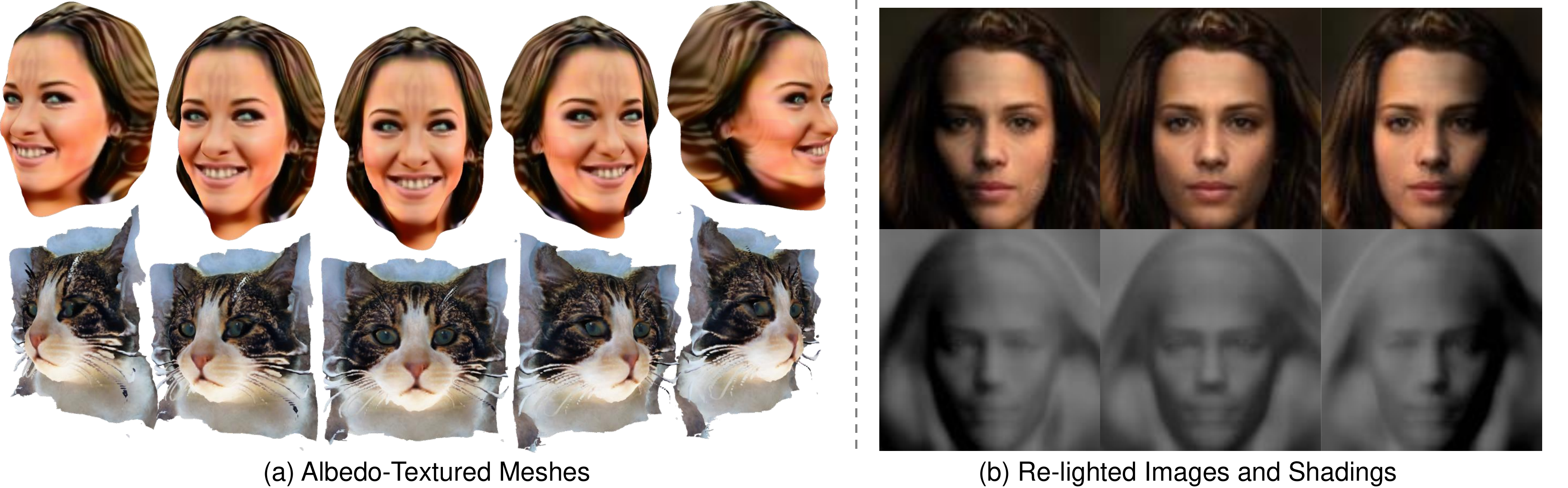}
    \vspace*{-0.7cm}
    \captionof{figure}{Generation results of \ourmodel. (a) indicates generated albedo-textured meshes and (b) presents re-lighted images and shadings.}
    \vspace*{-0.1cm}
\end{center}
}]

    

\begin{abstract}
\vspace{-0.3cm}
    While NeRF-based 3D-aware image generation methods enable viewpoint control, limitations still remain to be adopted to various 3D applications.
    Due to their view-dependent and light-entangled volume representation, the 3D geometry presents unrealistic quality, and the color should be re-rendered for every desired viewpoint.
    To broaden the 3D applicability from 3D-aware image generation to 3D-controllable object generation, we propose the factorized representations which are view-independent and light-disentangled, and training schemes with randomly sampled light conditions.
    We demonstrate the superiority of our method by visualizing factorized representations, re-lighted images, and albedo-textured meshes.
    Furthermore, we show that our approach improves the quality of the generated geometry via visualization and quantitative comparison.
    To the best of our knowledge, this is the first work that extracts albedo-textured meshes utilizing unposed 2D images without any additional labels or assumptions.
    
\end{abstract}


\vspace{-0.5cm}
\section{Introduction}
\label{sec:intro}
\vspace{-0.2cm}

Generating photo-realistic images has been a long-standing problem in computer vision.
Addressing this concern, generative adversarial networks (GANs)~\cite{gan} have shown their remarkable capability of generating high-quality images~\cite{dcgan,pggan,biggan,stylegan,stylegan2}. 
However, due to the lack of learning 3D features of an object, the previous GAN-based approaches have limitations on editing the content of a generated image in a 3D space (\eg viewpoint changing, re-lighting, etc.).
To enable controlling contents of the images in a 3D space, 3D-aware image synthesis~\cite{escaping,blockgan,hologan} has been proposed, generating a 3D representation of an object and then producing an image with a neural projection.
Among them, recent approaches~\cite{graf,pigan,giraffe} using neural radiance fields (NeRF)~\cite{nerf} presented a successful viewpoint control while preserving the multi-view consistency.
Particularly, Chan~\etal~\cite{pigan} showed a potential that the idea can be extended to 3D applications (\eg games and movies) by extracting the 3D mesh from the generated representation.

However, as also pointed out in previous work~\cite{pigan}, such approaches bear limitations as follows: 1) their texture representation is difficult to be rendered under various external environments (\ie view directions and light conditions) via a conventional computer graphics (CG) renderer, and 2) their 3D representations contain unrealistic geometry.

\begin{figure}[t!]
    \centering
    \includegraphics[width=1.0\linewidth]{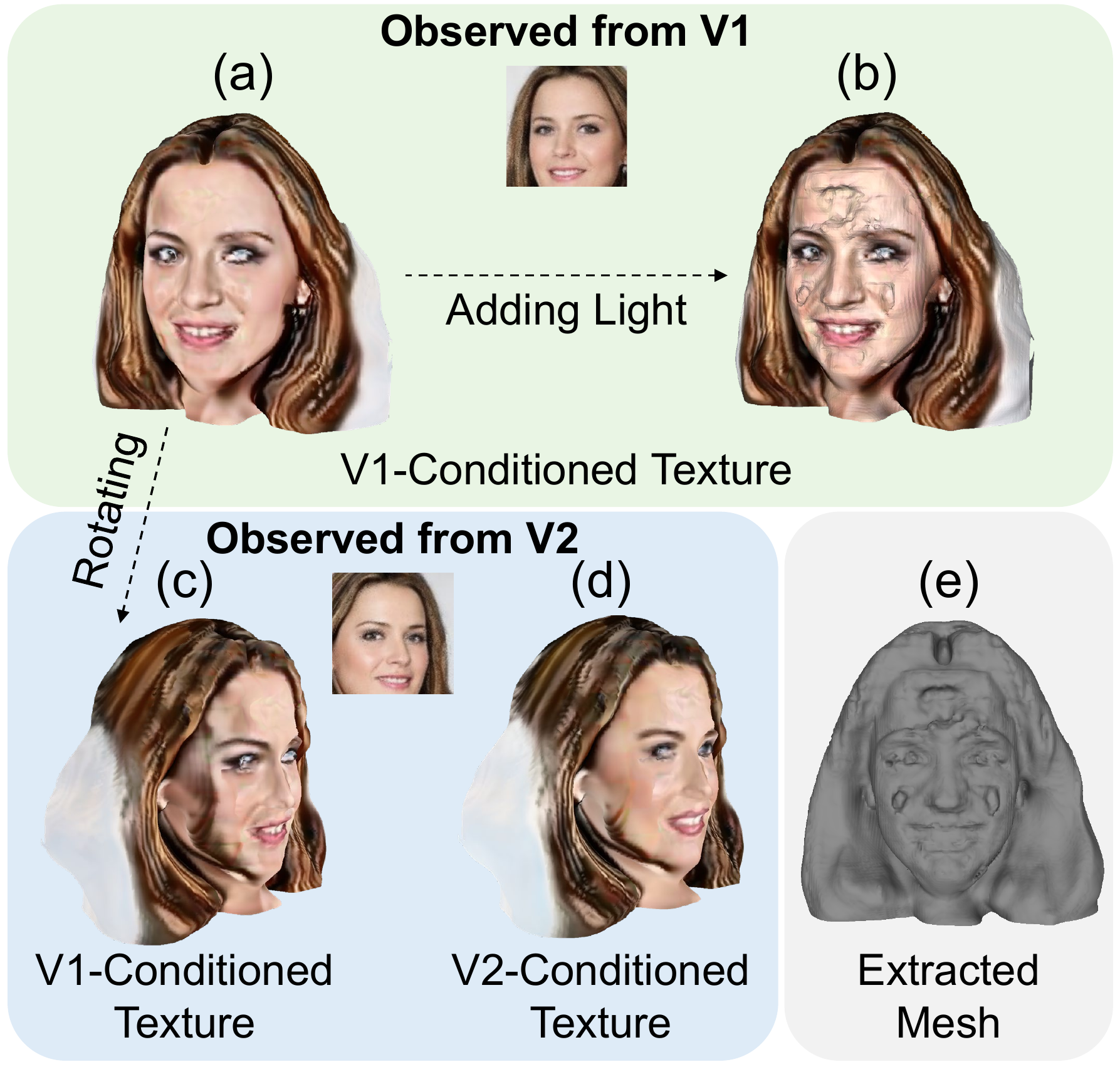}
    \vspace{-0.7cm}
    \caption{Generated textured mesh from pi-GAN~\cite{pigan}. (a) and (d) are the meshes observed from different view directions and textured with colors conditioned on their view directions (V1, V2).
    (c) shows the rotated mesh of (a), and (b) is a result of adding light on (a) via the conventional rendering pipeline. (e) describes the extracted geometry.}
    \label{fig:intro_failure}
    \vspace{-0.7cm}
\end{figure}

First, since they generate \emph{view-dependent} and \emph{light-entangled} 3D representations, the outputs are hardly applicable to the CG rendering pipeline that is necessary for various 3D applications.
The CG rendering process employs physical reflectance properties (\ie ambient, diffuse, and specular reflection) to render an object with diverse external environments.
Hence, we need an object with inherent feature representations (\eg albedo, normal, and specular coefficients), which are independent of view directions and light conditions; we call such an object a \emph{3D-controllable} object.
However, the previous approaches predict the color conditioned on the viewing direction to represent the plausible light reflection, which causes redundant the rendering problem when utilized in CG rendering pipelines.
Fig.~\ref{fig:intro_failure} illustrates textured meshes obtained from pi-GAN~\cite{pigan}, where the predicted colors are conditioned on two different view directions (V1 and V2).
As shown in Fig.~\ref{fig:intro_failure}(a), when we observe a mesh with V1-conditioned texture from the same direction (V1), it looks realistic.
However, (c) shows that the same mesh observed from a different view direction (V2) seems unrealistic.
To alleviate this issue, the colors have to be regenerated from the given view direction (V2), as presented in Fig.~\ref{fig:intro_failure}(d).
That is, we need to re-render an object for every desired view direction, which limits the applicability for 3D applications.
Also, since the light effects are already included in the generated colors, which is not albedo, adding different light conditions to an object produces unnatural visual quality.
To resolve these issues, we generate inherent 3D feature representations (\ie albedo, normal, and specular coefficients) without view dependency and explicitly compute the light reflection.
This allows our model to generate 3D-controllable objects beyond 3D-aware image generation.

Moreover, the existing methods have unrealistic surface geometry of their generated 3D representation due to the ambiguity in 3D-to-2D projection.
Light reflection (\eg diffuse and specular reflection) on the object depends on its surface geometry.
However, the previous methods predict colors only to compose a realistic 2D image at the given view direction without considering its geometry and light reflection.
This allows the models to generate an object with unrealistic geometry as presented in Fig.~\ref{fig:intro_failure}(e), while the projected 2D images look realistic (\ie ambiguity in 3D-to-2D projection).
Such defects can be manifested by simple light controls as described in Fig.~\ref{fig:intro_failure}(a) and (b), where (b) is the result of adding light to (a).
In other words, we can resolve the ambiguity by adding light reflection to the generated image.
Based on these insights, we force the geometry of generated 3D representation to be realistic by modeling the color from physical reflectance properties and explicitly controlling the light conditions while training.

In this paper, we propose \ourmodel that removes view and light dependency of the color representation and addresses 3D-to-2D projection ambiguity in a 3D-aware image generation task.
Addressing the limitations of the color representation of the previous work, our model generates \emph{view-independent} and \emph{light-disentangled} factorized representations (\ie albedo, normal, and specular coefficients) and renders the final image via photometric rendering (\ie shading model with light conditions and the factorized representations using physical reflectance properties).
Randomly sampling light conditions and deceiving a discriminator with rendered images, our model can generate multi-view realistic images with accurate geometry of their 3D representations.
Also, we introduce the staged training and the normal prediction layer to improve the training stability.
We demonstrate that \ourmodel successfully generates not only realistic multi-view 2D images but also factorized 3D representations with accurate geometry on various datasets.
Besides, since our model explicitly simulates the light reflection, it improves the 3D applicability of the generative model by producing re-lighted images and high-quality albedo-textured meshes.
To the best of our knowledge, this is the first work that generates 3D-controllable objects using unposed 2D images without any additional supervision.
We summarize our contributions as follows:
\begin{itemize}[noitemsep,topsep=0pt]
    \item We present \ourmodel that generates view-independent and light-disentangled volume representations (\ie albedo, normal, and specular coefficients), broadening its 3D applicability.
    \item We improve the geometry quality of generated 3D representations, resolving the 3D-to-2D projection ambiguity by modeling the color from physical reflectance properties.
    \item We validate the effectiveness of our model by presenting re-lighted images and albedo-textured meshes and providing qualitative and quantitative comparisons on the geometry quality of generated representations.
\end{itemize}

\section{Related Work}
\vspace{-0.2cm}
\label{sec:related}
\textbf{Neural 3D representations.}
Implicit representation has been employed by recent work~\cite{deepsdf, learningimplicit, localdeep, learningshape, occupancy, texturefields, occupancyflow, implicitsurface, pifu} for their memory efficiency and continuity, training the network to map 3D coordinates to occupancy field or signed distance functions.
Several methods~\cite{pifu, geopifu, pamir, neuralbody} extended this to texture representation and successfully extracted textured mesh from 2D images.
However, they still require 3D ground truth for training. 
To mitigate this undesirable necessity, differentiable rendering techniques are adopted to enable 2D image-based training~\cite{learningwo3d, scene, multiviewneural}.
Mildenhall~\etal~\cite{nerf} proposed neural radiance fields (NeRF) with volume rendering, showing impressive performance for novel-view synthesis.
To further improve this seminal work, some approaches~\cite{nerd, nerfactor, nerv} decomposed the view-dependent color of NeRF into reflectance properties and light conditions. NeRD~\cite{nerd} extracts albedo-textured meshes from the implicit 3D representations for real-time rendering and re-lighting.
However, the aforementioned approaches require multi-view and posed images for their training.
Therefore, we propose a novel generative architecture that only requires unposed single-view images for training.

\textbf{3D-aware image generation.}
Recent generative models utilized 3D representation to generate view-controllable images.
Several approaches employed voxel~\cite{escaping, hologan, blockgan} as their 3D representation.
Meanwhile, they generated coarse results without fine details due to the memory inefficiency of voxel representation.
Mitigating this issue, NeRF-based methods~\cite{graf, pigan} emerged, generating view-consistent 2D images in high quality.
Niemeyer~\etal~\cite{giraffe} extended this radiance field to feature field and adopted a 2D convolution neural renderer to further enhance the quality of generated images and rendering efficiency.
However, convolution operation on adjacent pixels hurts the internal 3D information, which leads to multi-view inconsistency and noisy geometry.
In addition, these NeRF-based methods output view-dependent and light-entangled colors, impeding them from being further extended to 3D object generation.

\vspace{0.1cm}
\textbf{3D object generation.}
Early 3D object generation methods required 3D supervision from voxel~\cite{3Dgan} and point cloud~\cite{learningimplicit, progressivepoint, 3Dpointgan}.
However, constructing 3D ground truth datasets is time-consuming and labor-intensive.
To avoid 3D supervision, several methods utilized 3D-morphable models~\cite{disco, stylerig} or 2D supervision~\cite{3dshape, pixel2mesh}.
Some of them~\cite{convmesh, meshvae} successfully generated textured mesh from 2D images with additional supervisions such as silhouette images and camera poses via differentiable rendering process.
Despite their success, they utilized ellipsoid template mesh as a starting point and still required additional supervisions.
These 3D objects can also be generated from the aforementioned NeRF-based models~\cite{graf, pigan} without requiring additional supervision other than 2D images.
Chan~\etal~\cite{pigan} presented generated meshed in its paper, showing a potential of applicability to 3D applications.
However, the generated meshes have a low quality of surface geometry, and even worse, the texture of mesh was difficult to be rendered with various external environments since it predicts view-dependent colors.
In this paper, we successfully address these concerns by factorizing view-dependent colors into inherent representations and explicitly controlling light conditions.
Table~\ref{Table:related_paper} shows our contributions by comparing the types of supervisions, assumptions, and goals of ours and previous 3D object generation methods.

\begin{table}[t!]
\centering
\small
    \scalebox{0.9}{
    \begin{tabular}{c|c|c|c}
    \toprule
    Paper & Supervision & Goals & Assumption\\
    \midrule
    \cite{3Dgan} & 3DV & 3DV (w/o T) &  \\
    \cite{3Dpointgan, progressivepoint} & 3DP & 3DP (w/o T) & \\
    \cite{learningimplicit} & 3DP & 3DM$^*$ (w/o T) & \\
    \midrule
    \cite{disco, stylerig} & I, 3DMM & MI, LI, 3DMM &  \\
    \cite{3dshape} & SI & 3DV (w/o T) & \\
    \cite{pixel2mesh} & I, KP & 3DM (w/o T) & Ellipsoid\\
    \cite{meshvae} & I, KP, BG, SI & 3DM & Ellipsoid\\
    \cite{convmesh} & I, SI, KP & 3DM & Ellipsoid\\
    \cite{ligtinggan} & I & MI, LI, A, D, N & Symmetry \\
    \midrule
    \cite{graf, pigan} & I & MI, 3DM$^*$ (w/o T) & \\
    Ours & I & MI, LI, 3DM$^*$, A, N, S & \\
    \bottomrule
    \end{tabular}
    }
    \vspace{-0.1cm}
    \caption{Comparison with the previous 3D object generation work. I: image, 3DV: 3D voxel, 3DP: 3D point cloud, 3DM: 3D mesh, 3DMM: 3D morphable model, KP: 2D keypoints or camera matrix, SI: silhouette image, BG: background image, MI: multi-view images, LI: re-lighting images, T: texture, A: albedo, N: normal, D: depth, S: specular coefficients. 3DM$^*$ denotes extracted 3D mesh from implicit representation via marching cube algorithm.}
    \label{Table:related_paper}
    \vspace{-0.7cm}
\end{table}


\vspace{-0.2cm}
\section{Proposed Method}
\label{sec:method}
\vspace{-0.1cm}
This section presents how we generate view-independent and light-disentangled volume representations for enhancing the 3D controllability of generated objects.

\begin{figure*}[t!]
    \centering
    \includegraphics[width=\linewidth]{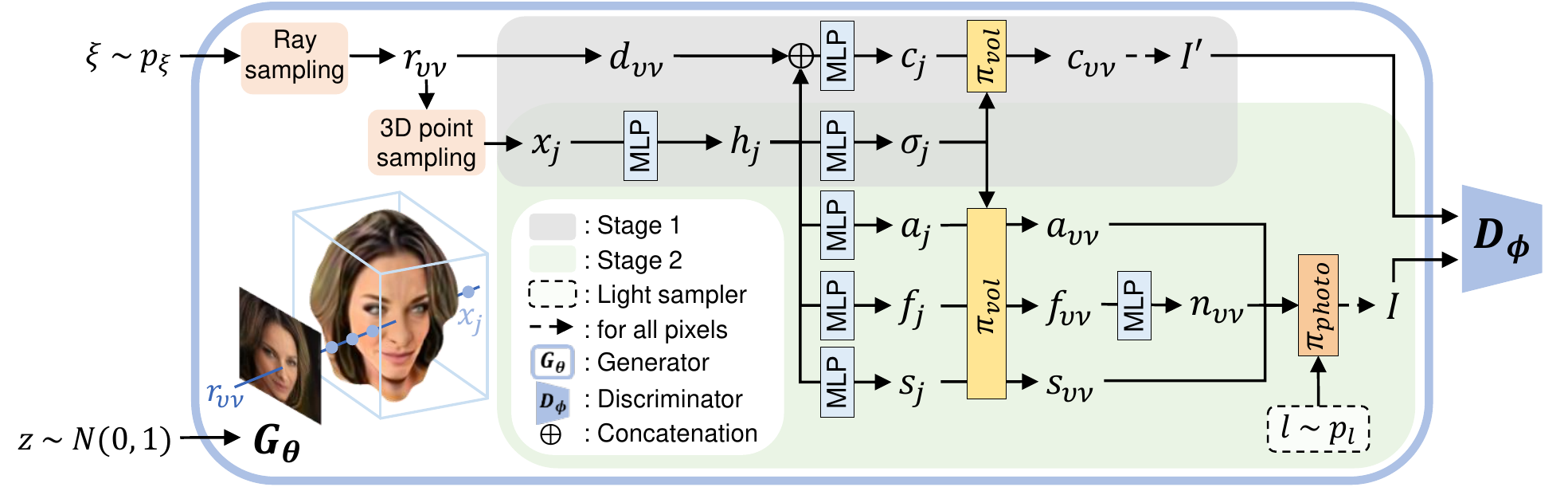}
    \vspace{-0.7cm}
    \caption{Overview of \ourmodel. To generate view-independent and light-disentangled colors, we disentangle the light and factorize the volume representations into albedo, normal, and specular coefficients. $\pi_{vol}$ and $\pi_{photo}$ denote volume rendering in Sec.~\ref{sec:preliminaries} and photometric image rendering in Sec.\ref{sec:architecture}, respectively. }
    \vspace{-0.3cm}
    \label{fig:overview}
\end{figure*}

\subsection{Preliminaries}\label{sec:preliminaries}
\textbf{Neural radiance field.}
Radiance field is a continuous function that maps a 3D location $\mathbf{x} \in \mathbb{R}^{3}$ and a viewing direction $\mathbf{d} \in \mathbb{S}^{2}$ to volume density $\sigma \in \mathbb{R}^{+}$ and view-dependent color $\mathbf{c} \in \mathbb{R}^{3}$. 
Inspired by NeRF~\cite{nerf}, recent approaches~\cite{graf, giraffe} use multi-layer perceptrons to represent this mapping function as
\begin{flalign}
\begin{split}
f_\theta: \mathbb{R}^{3} \times \mathbb{S}^{2} \rightarrow \mathbb{R}^{3} \times \mathbb{R}^{+}, \quad f_\theta (\mathbf{x}, \mathbf{d}) = (\mathbf{c}, \sigma).
\end{split}
\end{flalign}
To represent non-Lambertian effects of color, $\mathbf{d}$ is concatenated with intermediate feature vector of $f_{\theta}$ and passed to additional layers that output view-dependent RGB color $\mathbf{c}$.

\textbf{Volume rendering.}
With sampled camera pose $\xi$, ${\{\mathbf{x}_j\}^{N_x}_{j=1}}$ denotes sampled points along the camera ray for a pixel location $(u, v)\in\Omega$, where $\Omega$ indicates the set of pixel locations in the image plane, and $N_x$ denotes the number of sampled points in a camera ray.
To calculate the expected color $\mathbf{c}_{uv}$ for each pixel location $(u, v)$, the non-parametric volume rendering function $\pi_{vol}(\cdot)$ that maps the color $\mathbf{c}_j$ and the volume density $\sigma_j$ of the sampled point $\mathbf{x}_j$ to $\mathbf{c}_{uv}$ can be define as
\begin{equation}
\pi_{vol}: (\mathbb{R}^+, \mathbb{R}^3)^{N_x} \rightarrow \mathbb{R}^3, \quad \pi_{vol}\big(\{\sigma_j, \mathbf{c}_j\}^{N_x}_{j=1}\big) = \mathbf{c}_{uv}.
\end{equation}
To be specific, we obtain the estimated color $\mathbf{c}_{uv}$ via numerical integration~\cite{nerf} as
\begin{equation}
\begin{gathered}
    \vspace{-0.3cm}
    \mathbf{c}_{uv} = \sum_{j=1}^{N_x} w_j \mathbf{c}_j, \quad
    w_j = \tau_j \alpha_j, \\
    \tau_j = \prod_{k=1}^{j-1}(1-\alpha_k), \quad
    \alpha_j = 1 - e^{-\sigma_j \delta_j},
    \vspace{-0.1cm}
    \label{eq:volume_rendering}
\end{gathered}
\end{equation}
where $\tau_j, \alpha_j$ denote the transmittance and the alpha value for $\mathbf{x}_j$, and $\delta_j=||\mathbf{x}_{j+1} - \mathbf{x}_j||_2$ is the L2-distance between the adjacent sampled points.

To this end, we extend this definition to the volume parameter $\mathbf{p}$, similar to Boss~\etal~\cite{nerd}, which can be one of our factorized volume representations such as albedo $\mathbf{a}$, intermediate feature vector $\mathbf{f}$, and specular coefficients $\mathbf{s}$.
These factorized representations obtained from our network are volume rendered as 
\begin{equation}
\begin{gathered}
    \vspace{-0.3cm}
    \mathbf{p}_{uv} = \sum_{j=1}^{N_x} w_j \mathbf{p}_j, \quad
    w_j = \tau_j \alpha_j. \\
    \vspace{-0.1cm}
    \label{eq:volume_rendering}
\end{gathered}
\end{equation}
\textbf{Generative NeRF.}
Schwarz~\etal~\cite{graf} proposed a NeRF-based generative method that conditions the model on the latent code $\mathbf{z} \in \mathbb{R}^{M}$, formulated as
\begin{flalign}
\begin{split}
g_\theta: \mathbb{R}^{3} \times \mathbb{S}^{2} \times \mathbb{R}^{M} \rightarrow \mathbb{R}^{3} \times \mathbb{R}^{+}, \quad g_\theta (\mathbf{x}, \mathbf{d}, \mathbf{z}) = (\mathbf{c}, \sigma).
\end{split}
\end{flalign}
The model generates implicit radiance field that maps a 3D location and a view direction to density and color conditioned on randomly-sampled latent vector $\mathbf{z}$.
The generator is trained to deceive the discriminator with volume-rendered 2D images under a sampled viewpoint. 

\subsection{Generative factorized volume representations}\label{sec:architecture} 
To overcome the aforementioned limitations of the previous methods in Sec.~\ref{sec:intro}, we propose \ourmodel that aims to generate view-independent and light-disentangled volume representations.
We factorize the representations into albedo, normal, and specular coefficients to obtain the final texture by leveraging physical reflectance properties.

\textbf{Factorized volume representations.}
As shown in Fig.~\ref{fig:overview}, we re-parameterize the view-dependent color of the previous methods~\cite{nerf, graf, pigan} with normal $\mathbf{n}\in\mathbb{R}^3$, specular coefficients $\mathbf{s}\in\mathbb{R}^2$, and view-independent color $\mathbf{a}\in\mathbb{R}^3$ that is interpreted as albedo. 
The factorized volume representations of our model are predicted as
\begin{equation}
\begin{split} 
h_\theta: \mathbb{R}^{3} \times \mathbb{R}^{M} \rightarrow \mathbb{R}^{H},& \quad \quad h_\theta (\mathbf{x}_j, \mathbf{z}) = \mathbf{h}_j \\
\sigma_\theta: \mathbb{R}^{H} \rightarrow \mathbb{R}^{+},& \quad \quad \quad \sigma_\theta (\mathbf{h}_j) = \sigma_j \\
a_\theta: \mathbb{R}^{H} \rightarrow \mathbb{R}^{3},& \quad \quad \quad a_\theta (\mathbf{h}_j) = \mathbf{a}_j\\
f_\theta: \mathbb{R}^{H} \rightarrow \mathbb{R}^{4},& \quad \quad \quad f_\theta (\mathbf{h}_j) = \mathbf{f}_j \\
s_\theta: \mathbb{R}^{H} \rightarrow \mathbb{R}^{2},& \quad \quad \quad s_\theta (\mathbf{h}_j) = \mathbf{s}_j, 
\end{split}
\end{equation}
where $\mathbf{h}_j$ is an intermediate feature vector conditioned on $\mathbf{z}$. 
The prediction layers $\sigma_{\theta}$, $a_{\theta}$, $f_{\theta}$, and $s_{\theta}$ are implemented with a linear layer.
We obtain the normal after volume rendering the predicted normal features $\mathbf{f}$ and passing it through additional linear layers as
\begin{flalign}
\begin{split}
n_\theta: \mathbb{R}^{4} \rightarrow \mathbb{R}^{3}, \quad \quad \quad n_\theta (\mathbf{f}_{uv}) = \mathbf{n}_{uv},
\end{split}
\end{flalign}
where $n_{\theta}$ consists of two linear layers and normalization to a unit vector.


Although the normal of the implicit function at a point $j$ can be obtained by  $\frac{-\nabla_\textbf{x}\sigma_j}{\|\nabla_\textbf{x}\sigma_j\|}$~\cite{nerd, nerfactor, multiview}, utilizing this directly for the photometric rendering can lead to sub-optimal results for the following reasons.
First, as also pointed out in Zhang~\etal~\cite{nerfactor}, the gradient of volume density tends to be noisy.
This makes it difficult to render clear images when exploiting the gradient of $\sigma$ for rendering.
Moreover, the second-order gradient has to be computed for optimization, which is computationally demanding.
To address such concerns for stabilizing the training procedure, we introduce an additional normal prediction layer and use the output (\ie predicted normal) as a proxy of the gradient of $\sigma$.
Afterward, we devise a regularization term that utilizes the predicted normal as weak supervision for the gradient of $\sigma$, stabilizing the optimization process.
Our final normal loss is defined as
\begin{flalign}
\begin{split}
    \vspace{-0.3cm}
    &\mathcal{L}_\text{normal} =\\
    &\frac{1}{|\Omega|}\sum_{(u,v) \in \Omega} \bigg\{
    || \mathbf{n}_{uv} - \sum_{j=1}^{N_x} w_j \frac{-\nabla_\mathbf{x}\sigma_j}{\|\nabla_\mathbf{x}\sigma_j\|_2}||_2 
    + \lambda \mathcal{L}_\text{TV}\left(\mathbf{n}_{uv}\right)
    \bigg\},
\end{split}
\label{normalloss}
\vspace{-0.3cm}
\end{flalign}
where $\lambda$ denotes the hyper-parameter indicating the relative importance between two losses, and $\mathcal{L}_{TV}$ represents the difference between the center pixel value and its surrounding ones for smoothing the normal in the local region.

\textbf{Photometric image rendering.}
The final textured image is rendered from a combination of albedo, normal, specular coefficients, and external environment via photometric rendering function $\pi_{photo}(\cdot)$.
We sample light condition $l=(k_a, k_d, l_d)$ from the prior distribution, where $l$ consists of the intensity of ambient light $k_a\in\mathbb{R}^{+}$, the intensity of directional light $k_d \in \mathbb{R}^{+}$, and the direction of light $l_d \in \mathbb{R}^3$.
We calculate the dot product of the normal $\mathbf{n}_{uv}$ and light direction $l_d$ to obtain a value for the directional illumination.
Afterward, the shading map $\mathbf{H}$ is obtained by multiplying the intensity of directional light and adding the intensity of ambient light to the obtained value: $\mathbf{H}_{uv} = (k_a + k_d \cdot max\{0, \langle l_d, \mathbf{n}_{uv} \rangle \})$.
The specular map $\mathbf{S}$ is obtained from the predicted specular coefficients, normal, directional light and view direction:  $\mathbf{S}_{uv} = k_d \cdot \mathbf{s}^{(0)}_{uv} \cdot (max\{0, \cos \theta\})^{\mathbf{s}^{(1)}_{uv}}$, where $\theta$ denotes the angle between the view direction and the light direction after reflected by normal, and $\mathbf{s}^{(0)}, \mathbf{s}^{(1)}$ denote the coefficient for the specular reflection and shininess, respectively.
Note that this process is calculated for each pixel $(u, v)$ in the image plane.
Finally, we render the final textured image from a combination of albedo, shading map, and specular map, formulated as
\begin{equation}
\mathbf{I} = \mathbf{a} \cdot \mathbf{H} + \mathbf{S}.
\end{equation}

\begin{figure*}[ht]
    \centering
    \includegraphics[width=\linewidth]{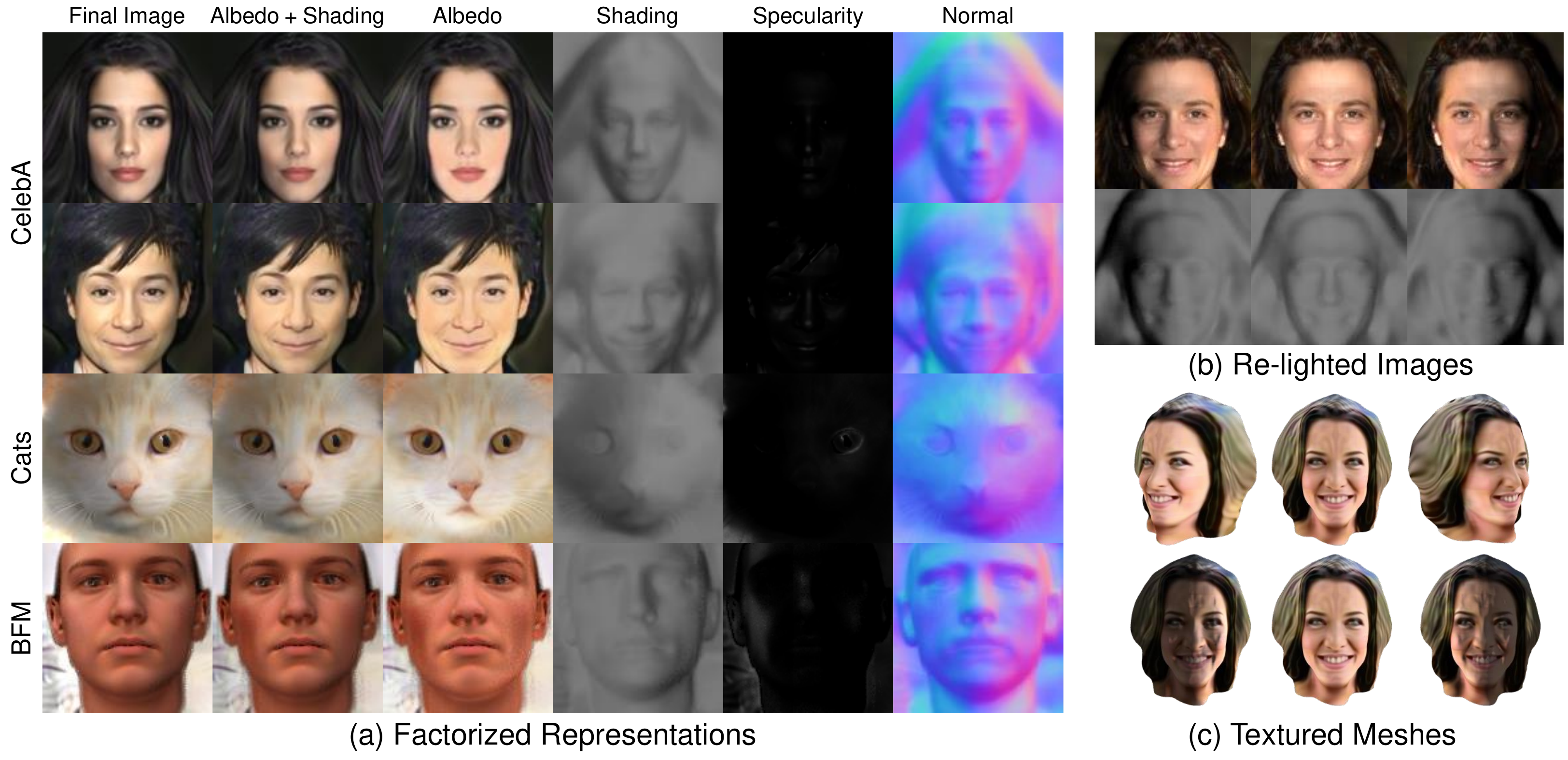} 
    \vspace*{-0.6cm}
    \caption{Visualization of the (a) factorized representations, (b) re-lighted images, and (c) textured meshes. Our method generates high-quality factorized representations and conducts re-lighting. The albedo-textured mesh generated from our method can be adopted to the conventional rendering pipelines.
    Note that we train the model without disentangling specularity for BFM dataset.}
    \vspace*{-0.5cm}
    \label{fig:exp_disentangle}
\end{figure*}

\subsection{Training}
\label{sec:procedure}

\textbf{Training objective function.}
Fig.~\ref{fig:overview} depicts the overall training procedure.
Our model first samples a latent variable $\mathbf{z} \sim \mathcal{N}(0, \mathbf{I})$ to condition the model.
We sample points $\{\textbf{x}_{j=1}\}_j^{N_x}$ on the ray direction from a random camera pose $\xi \sim p_\xi$ and sample the random light condition $l \sim p_l$. 
Then, we train our model $G_\theta$ using the non-saturating GAN~\cite{gan} loss with R1 regularization~\cite{r1_regularization}:
\begin{equation}
\begin{split}
    \vspace{-0.3cm}
    \mathcal{L}_\text{GAN}(\theta, \phi) = \mathbf{E}_{\mathbf{z} \sim p_z, \xi \sim p_\xi, l \sim p_l}[f(D_{\phi}(G_{\theta}(\mathbf{z}, \xi, l)))] \\
    + \mathbf{E}_{\mathbf{I} \sim  p_D}[f(-D_{\phi}(\mathbf{I})) 
    + \lambda \|\nabla D_{\phi}(\mathbf{I})\|^{2}],
    \vspace{-0.3cm}
\label{GANloss}
\end{split}
\end{equation}
where $f(x) = -log(1 + exp(-x))$, $p_D$ denotes the real data distribution, and $D_{\phi}$ indicates the discriminator.
$D_{\phi}$ is trained by maximizing Eq.~\eqref{GANloss}, and the generator is trained by minimizing both Eq.~\eqref{normalloss} and Eq.~\eqref{GANloss}: 
\begin{equation}
    \mathcal{L}_{total} = \mathcal{L}_{GAN} + \lambda_{n}\mathcal{L}_{normal},
\end{equation}
where $\lambda_{n}$ denotes the hyper-parameter indicating the relative importance between two losses.

\textbf{Staged training.}
To initialize the geometry of volume representation, we first train our network to generate the view-dependent color image $\mathbf{I}^{\prime}$ similar to the previous work~\cite{graf, pigan}.
Next, for the second stage, we attach the newly proposed layers for $(\mathbf{a}, \mathbf{n}, \mathbf{s})$ to the generator, and fine-tune the whole model to generate the images $\mathbf{I}$ from photometric rendering.
Utilizing the volume representation that has roughly predicted geometry as the initial point, we can stabilize the training of the factorized representations.

\subsection{Implementation details}\label{sec:details}
For our first stage training, we adopt the FiLMed SIREN architecture of Chan~\etal~\cite{pigan}, which leverages a mapping network for conditioning $\mathbf{z}$ and sinusoidal activation for improving image quality.
Regarding newly attached layers in the second stage, we use the different activation functions for the linear layers that map the intermediate feature vectors to our factorized representations: \{$\sigma_{\theta}, s_{\theta}$ : ReLU\}, \{$a_{\theta}$, $f_{\theta}$ : Sigmoid\}, and \{$n_{\theta}$ : Tanh\}.
We reduce the batch size to half for the second stage training.
Further implementation details are presented in supplementary materials.
\section{Experiments}
\label{sec:exp}

\subsection{Experiment setup}
\textbf{Datasets.}
We conduct experiments on two real-world datasets, CelebA~\cite{celeba} and CATs~\cite{cats}, and a synthetic dataset, BFM (Basel Face Model)~\cite{bfm}. 
CelebA dataset contains more than 200K face images of 10,177 unique identities. We utilize the aligned and cropped images with the resolution of 128$\times$128.
CATs dataset consists of over 6K in-the-wild cat images. We align and crop the cat faces based on the given annotations indicating the head of cats.
Additionally, we utilize a synthetic face model, BFM dataset, to evaluate the quality of geometry with the ground truth depth.
For all datasets, the images are resized to 64$\times$64 during the training, and we generate images at the resolution of 128$\times$128 and volume representations at 128$\times$128$\times$128 for mesh extraction.
In the training, we only utilize the unposed 2D images without any additional labels or assumptions.

\textbf{Baselines.}
As our baselines, we adopt the previous NeRF based 3D-aware image synthesis models: GRAF~\cite{graf}, pi-GAN~\cite{pigan}, and GIRAFFE~\cite{giraffe}.
Note that GIRAFFE involves convolution operations for their neural rendering process, thus, we do not extract meshes from this model.
We implement and train each model following the official codes.

\subsection{Interpreting the factorized representations}
\textbf{Disentanglement.}
We present the qualitative results of factorized representations and light disentanglement in Fig.~\ref{fig:exp_disentangle}.
As shown in Fig.~\ref{fig:exp_disentangle}(a), ours generates reasonable albedo, shading, specularity, and normal images.
Surprisingly, even though we do not provide the model with additional supervision, ours successfully achieves factorization.
Combining those factorized representations, our model outputs realistic final images.
Note that predicted specularity makes the final images more natural.

Furthermore, from these factorized representations, we successfully re-light 2D images under various light conditions, as illustrated in Fig.~\ref{fig:exp_disentangle}(b).
One can observe the shading image effectively reflects the given light conditions.

\textbf{Textured mesh.}
Thanks to our view-independent and light-disentangled representations, we successfully generate high-quality albedo-textured meshes.
Fig.~\ref{fig:exp_disentangle}(c) presents the meshes controlled via a conventional rendering pipeline.
The rendered meshes look realistic in various viewpoints and under diverse light conditions.
This is because the meshes are textured with albedo where all the effects of external environments are removed.






\begin{figure}[t!]
    \centering
    \includegraphics[width=\linewidth]{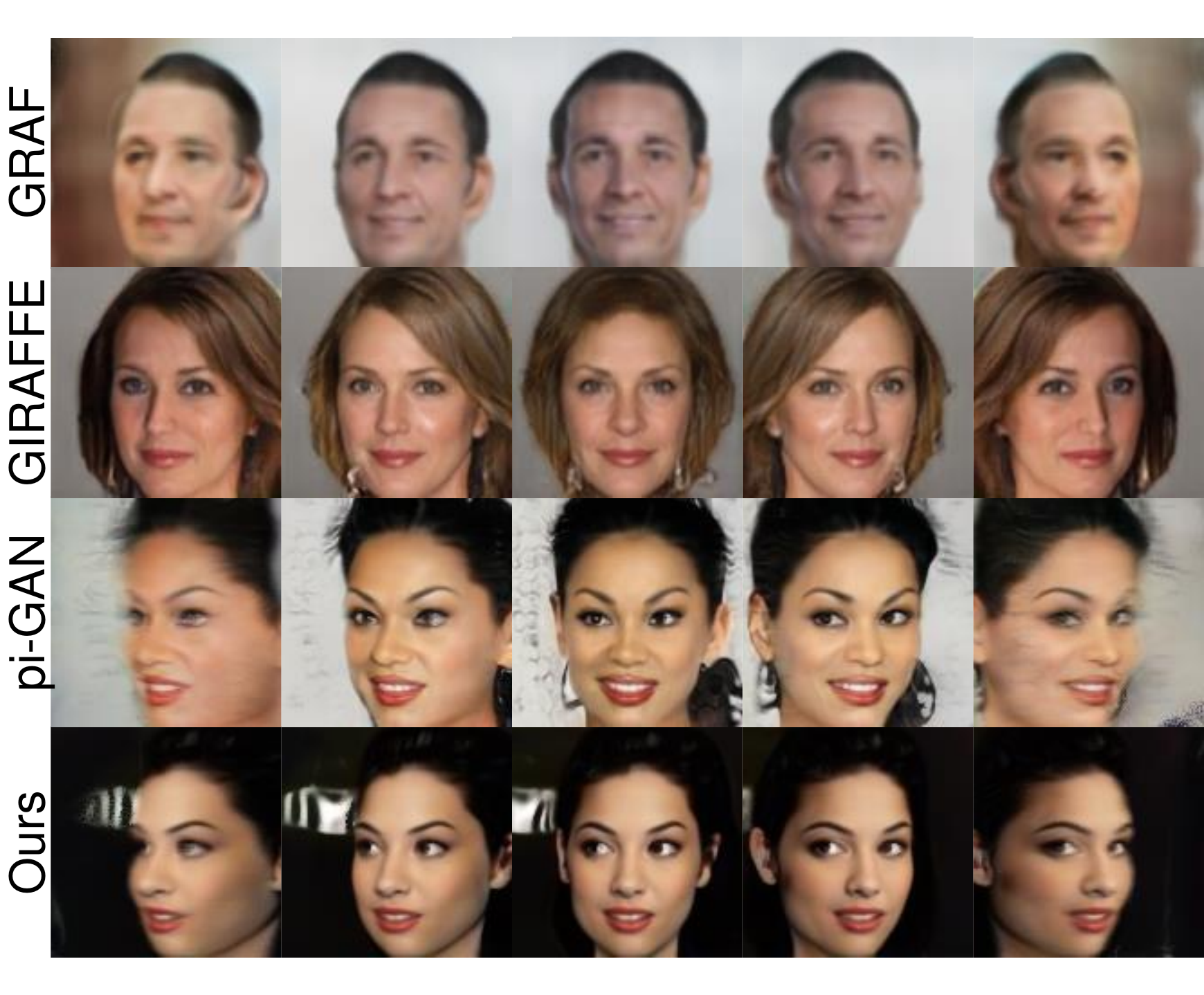}
    \vspace{-0.7cm}
    \caption{Comparison of multi-view images with baselines on CelebA and BFM dataset. Our method preserves the multi-view consistency while other baselines fail in extreme cases.}
    \vspace{-0.2cm}
    \label{fig:exp_multiview}
\end{figure}


\subsection{Comparison with baselines}

\textbf{Metrics.}
For the quantitative comparison, we show 1) the Frechet inception distance (FID) score for the image quality, 2) scale-invariant depth error (SIDE) for the geometry quality, and 3) projection FID for the quality of the textured mesh.
To measure SIDE~\cite{side}, we obtain depth maps of the generated images by volume rendering the distance between the given camera position and sampled points, as done in the previous approaches.
Moreover, for the ground truth depth of the generated images, we use the depth estimation model~\cite{depth} pre-trained with BFM dataset.
Projection FID score is the FID score of images projected from the generated textured meshes via the conventional rendering pipeline.
To obtain textured meshes, we apply a marching cube algorithm on volume density to extract meshes and use albedo at surface points as texture.
For the baselines, which require a view direction to predict the color, we use the color from the frontal view direction to obtain the texture.
Afterward, we project the textured meshes into images with randomly-sampled viewpoints using Blender~\cite{blender} and then measure the projection FID score.
Further details are presented in supplementary materials.

\begin{table}[t!]
\centering
\small
    \scalebox{0.95}{
    \begin{tabular}{c|c|c|c|cc}
    \toprule
    \multirow{2}{*}{\shortstack{View\\dependency}} & \multirow{2}{*}{Model} & CelebA & CATs & \multicolumn{2}{c}{BFM}\\
    & & FID$_{\downarrow}$ & FID$_{\downarrow}$ & FID$_{\downarrow}$ & SIDE$_{\downarrow}$\\
    \midrule
    \multirow{3}{*}{\ding{51}}& GRAF & 38.86  & 26.12 & 56.11 & 3.19 \\
    &pi-GAN & \textbf{14.25} & \textbf{10.77} & \textbf{14.83} & 1.78 \\
    &GIRAFFE & 22.02 & 21.01 & 18.46 &  - \\
    \midrule
    \multirow{2}{*}{\ding{55}}&pi-GAN* & \textit{18.10} & 15.77 & 21.54 & - \\
    &Ours & 20.17  & \textit{13.56} & \textit{19.94} & \textbf{1.41} \\
    \bottomrule
    
    \end{tabular}
    }
    \vspace{-0.1cm}
    \caption{Quantitative evaluations with the FID and SIDE scores using CelebA, CATs, and BFM dataset. pi-GAN* denotes a variant where the view-dependency of color representation is removed. Note that SIDE score is scaled by $10^2$ for better visualization.}
    \label{Table:exp_baseline_comparison}
    \vspace{-0.2cm}
\end{table}

\begin{figure}[t!]
    \centering
    \includegraphics[width=\linewidth]{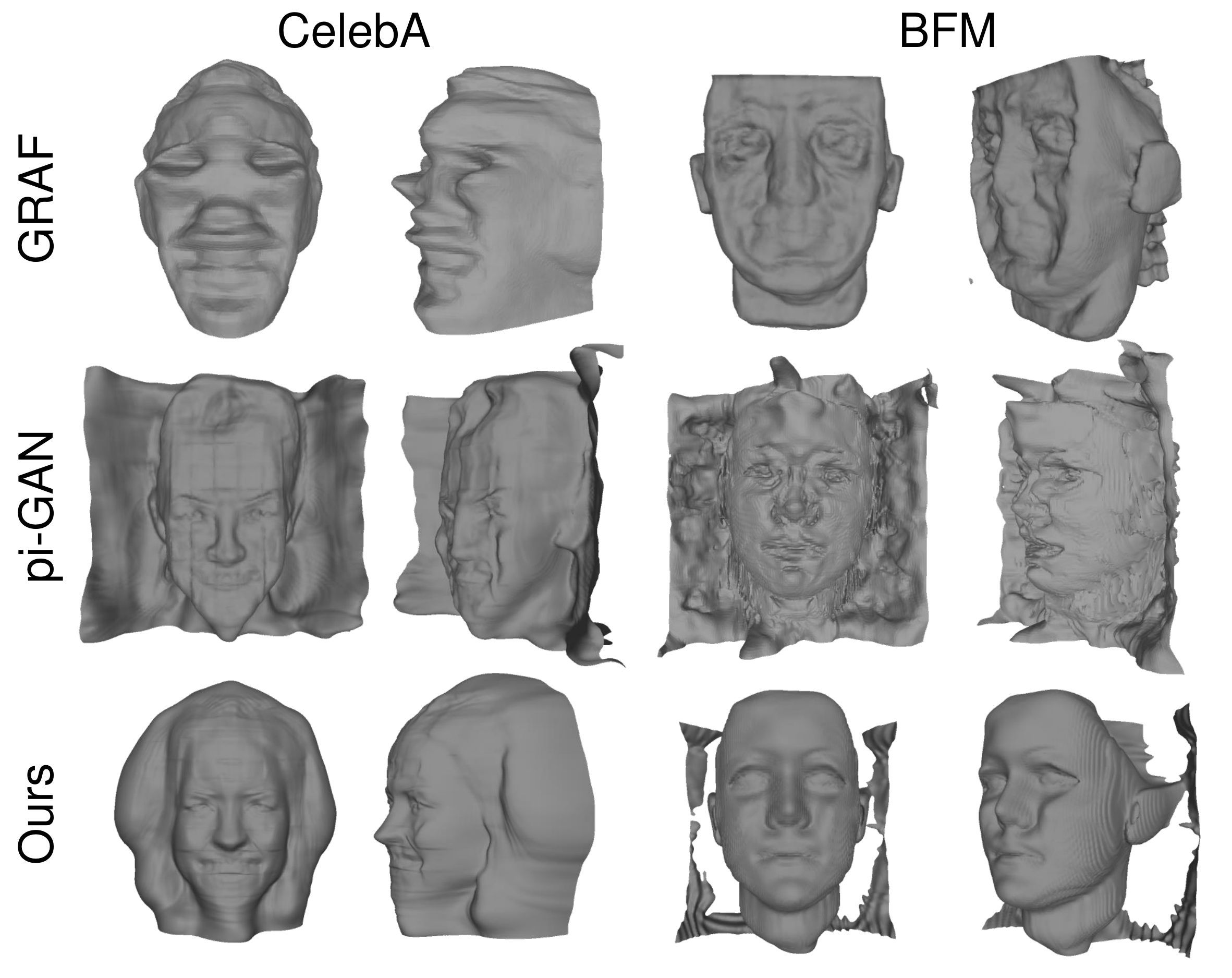} 
    \vspace{-0.5cm}
    \caption{Comparison of the extracted meshes with baselines on CelebA and BFM dataset. Unlike other baselines, we can observe that our model generates more convex meshes.}
    \vspace{-0.5cm}
    \label{fig:geometry}
\end{figure}

\textbf{Quality of generated images}
To evaluate the quality of generated images, we measure the FID scores on CelebA, CATs, and BFM dataset as illustrated in Table~\ref{Table:exp_baseline_comparison}.
Our method achieves reasonable FID scores, comparable to other baselines, involving marginal performance drop.
We conjecture that this is mainly due to the view independency of our representations.
The view dependent color representation of previous work effectively accounts for dataset bias (\eg eye movement toward the front) and lowers the FID scores.
For this reason, the variant of pi-GAN (\ie pi-GAN*) where the view-dependency of color representation is removed shows performance drop and the similar FID scores with the ones of ours.
That is, our proposed approach has a comparable capability to pi-GAN in generating high-quality images.
\begin{table}[t!]
\centering
\small
    \scalebox{0.78}{
    \begin{tabular}{p{0.072\textwidth}<{\centering}|
    p{0.018\textwidth}<{\centering}|p{0.018\textwidth}<{\centering}|
    p{0.018\textwidth}<{\centering}|p{0.018\textwidth}<{\centering}|
    p{0.018\textwidth}<{\centering}|p{0.018\textwidth}<{\centering}| 
    p{0.018\textwidth}<{\centering}|p{0.018\textwidth}<{\centering}|
    p{0.018\textwidth}<{\centering}|
    p{0.018\textwidth}<{\centering}|p{0.018\textwidth}<{\centering}|
    p{0.018\textwidth}<{\centering}}
    \toprule
    & \multicolumn{6}{c|}{pi-GAN} & \multicolumn{6}{c}{Ours} \\
    \midrule
    Thresholds    & 60 & 70 & 80 & 90          & 100 & 110    & 20 & 30          & 40 & 50 & 60 & 70\\
    Proj. FID      & 56 & 55 & 55 & \textbf{54} & 55  & 62     & 53 & \textbf{50} & 51 & 52 & 53 & 54 \\
    \midrule
    Best & \multicolumn{6}{c|}{53.84} & \multicolumn{6}{c}{\textbf{50.17}} \\
    \bottomrule
    \end{tabular}
    }
    \vspace{-0.1cm}
    \caption{Projection FID scores of ours and baselines on CelebA dataset along with the sigma threshold of the marching cube algorithm.}
    \label{Table:exp_projection}
    \vspace{-0.3cm}
\end{table}
\begin{figure}[t!]
    \centering
    \includegraphics[width=\linewidth]{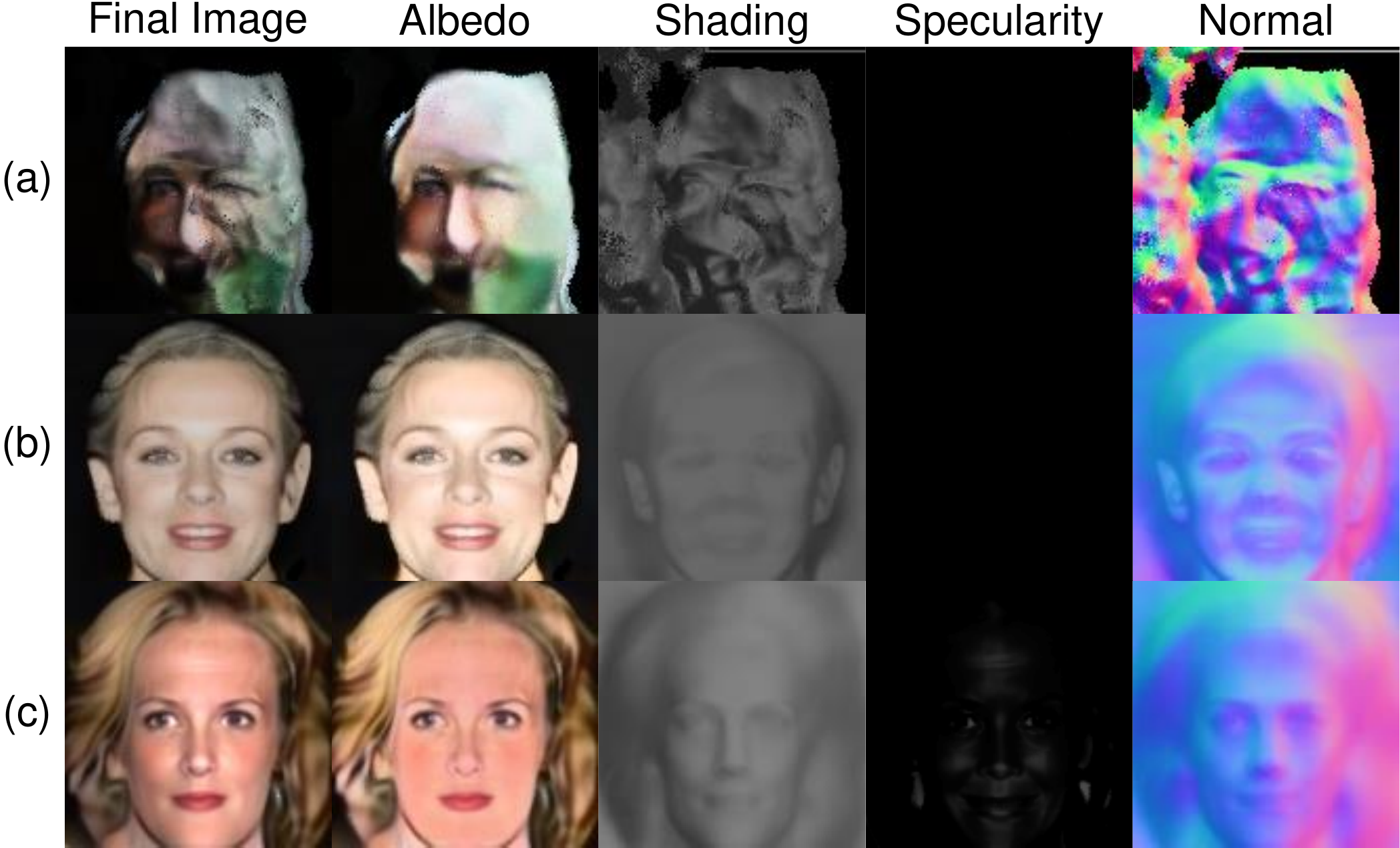} 
    \vspace{-0.7cm}
    \caption{Visualization of factorized representations for ablation study. Images in (a) are from the model trained without normal prediction layer.
    (b) indicates generated images from our model without specularity disentanglement.
    The last row (c) is the results of our final model.}
    \vspace{-0.5cm}
    \label{fig:exp_ablation}
\end{figure}

\begin{figure*}[t!]
    \centering
    \includegraphics[width=\linewidth]{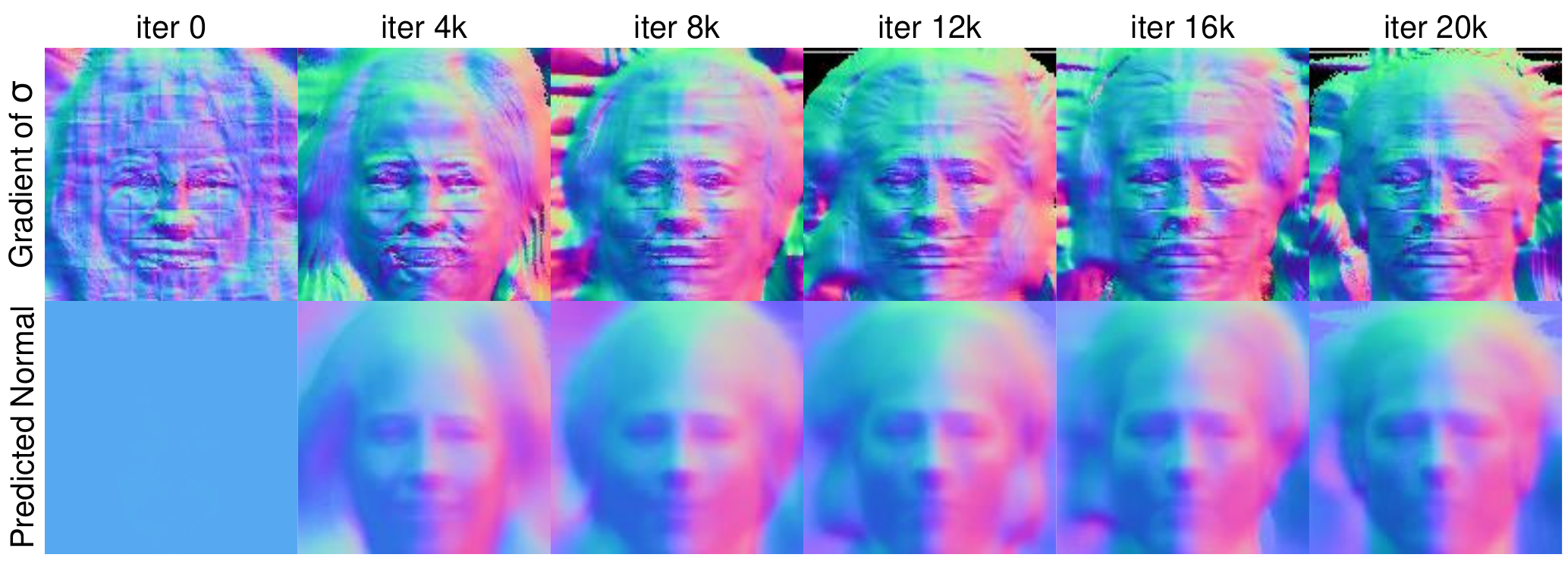}
    \vspace{-0.7cm}
    \caption{Normal refinements over iteration changes. As shown, gradient of $\sigma$ gets gradually refined over training. In addition, predicted normal converges fast and works as a guidance to form better geometry.}
    \vspace{-0.5cm}
    \label{fig:nomal_refine}
\end{figure*}
\textbf{Multi-view consistency.}
We compare the quality of the generated multi-view images from ours and baselines.
As Fig.~\ref{fig:exp_multiview} describes, our method preserves reasonable multi-view consistency even in extreme viewpoints while other baselines generally fail.
Since the baselines require view direction to predict the color, they fail in generating color from the extreme viewpoints, which is scarce in training datasets.
Meanwhile, our model predicts view-independent color, resulting in more reasonable image generation quality from extreme viewpoints.
Also, in the case of GIRAFFE~\cite{giraffe}, multi-view inconsistency becomes even worse (\eg hair direction changes) since GIRAFFE~\cite{giraffe} employs a 2D neural renderer composed of convolution operations, which hurts the internal 3D-aware feature.


\textbf{Quality of 3D representation.}
To evaluate the quality of the generated 3D representations, we measure the depth error (SIDE) and the projection FID score of textured meshes.
As presented in Table~\ref{Table:exp_baseline_comparison}, our method outperforms baselines with a large gap in the depth error.
That is, the quality of geometry are improved by applying our method, and this can also be shown in Fig.~\ref{fig:geometry}.
Regarding the projection FID scores, while our model shows the lower FID scores than pi-GAN, the improvements look marginal in Table~\ref{Table:exp_projection}.
However, this is because the generated textured meshes from pi-GAN can seem realistic in frontal view, which is prevalent in the projected images since we sample viewpoints from Gaussian distribution.
As shown in Fig.~\ref{fig:geometry} and the Fig.~\ref{fig:projection_fid} of supplementary, the dents of the geometry is not easily identified in the frontal view, unlike in other view.

\subsection{Ablation studies}
Fig.~\ref{fig:exp_ablation} demonstrates the importance of each component of our model.
Fig.~\ref{fig:exp_ablation}(a) indicate the results of our model trained with the gradient of $\sigma$ as normal, and (b) illustrates the results of our model trained with predicted normal (\ie the output of normal prediction layer) and without specularity disentanglement.
Fig.~\ref{fig:exp_ablation}(c) are the results of our final model.
As shown in Fig.~\ref{fig:exp_ablation}(a) and described in Sec.~\ref{sec:method}, the training is unstable if we use the gradient of $\sigma$ as normal.
On the other hand, as presented in Fig.~\ref{fig:exp_ablation}(b), the training becomes stable if we utilize the predicted normal instead of the gradient of $\sigma$.
However, specularity still remains in albedo in (b), since we do not separate the specularity here.
Finally, our full model generates valid factorized representations as illustrated in Fig.~\ref{fig:exp_ablation}(c).

As described in Sec~\ref{sec:architecture}, the predicted normal from the normal prediction layer acts as weak supervision on the gradient of $\sigma$.
This results in the improvement of the quality of geometry of the generated volume representations.
We demonstrate the effectiveness of the normal prediction layer by visualizing the refinement progress of the gradient of $\sigma$ and the predicted normal during Stage 2 training, as illustrated in Fig.~\ref{fig:nomal_refine}.
As the training proceeds, the predicted normal converges fast and presents more reasonable shape, and the gradient of $\sigma$ gets refined accordingly.





\section{Discussion}


\textbf{Limitations.}
First, we assume a single directional light for our method, which may not be appropriate for datasets containing complex light conditions.
Next, while normal is defined on the surface of an object, we utilize volume rendering process to calculate normal.
Although our model achieves successful results, focusing on surface to obtain normal could be one of the future work.
As another trial, we attempted to calculate the reflection at each sampled 3D point and apply volume rendering on this reflected light, instead of volume rendering normal and calculating the reflection in images. 
Though this seems more technically sound, the quality of the generated geometry and images degraded. 


\textbf{Ethical considerations.} 
Since we train ours on publicly available datasets, there may exist underlying bias in those datasets. 
This can result in lacking diversity of our generated images and 3D objects.
Moreover, 3D representations can be obtained with inverse rendering techniques, the method can be abused for generating DeepFakes. 
We are aware of the fact that DeepFake is a substantial threat in our society, and we never approve of employing our model for DeepFake generation with inappropriate intentions.

\section{Conclusion}
We propose \ourmodel that generates 3D-controllable objects beyond 3D-aware image generation.
By combining physical reflectance prior with the model, we generate an object with factorized representations that are independent of external environments. 
With our method, we can generate 3D-controllable objects that can be adopted in 3D conventional rendering pipelines and have realistic geometry.
We believe our work bridges a gap between content generation techniques and real-world 3D applications.


{\small
\bibliographystyle{ieee_fullname}
\bibliography{egbib}
}

\clearpage

\def\thesection{\Alph{section}}
\setcounter{section}{0}

\section{Overview}
In this supplementary material, we provide further implementation and experiment details.
Afterward, we illustrate additional experiment results and analyses.
Finally, we introduce a concurrent work, ShadeGAN, and present comparisons with them.
Furthermore, we present qualitative results on viewpoint and light condition control with videos, attached to supplementary materials.

\section{Implementation Details}
\label{sec:implementation}

\subsection{Architecture Details}
\textbf{FiLMed SIREN network.}
We adopt FiLMed SIREN network proposed by Chan~\etal~\cite{pigan} as our backbone, which leverages a mapping network conditioning and sinusoidal activations.
The FiLMed SIREN backbone is formulated as
\begin{equation}
\begin{split}
    h_\theta(\mathbf{x}_j) = \phi_{n-1} \circ \phi_{n-2} \circ \mathellipsis \circ \phi_{0} ~ (\mathbf{x}_j), \\
    \phi_{i}(\mathbf{x}_j^{(i)}) = \sin (\boldsymbol{\gamma}_i \cdot (\mathbf{W}_i \mathbf{x}^{(i)}_j + \mathbf{b}_i) + \boldsymbol{\beta}_i),
        \label{eq:pigan_1}
\end{split}
\end{equation}
where $\mathbf{x}_j$ indicates the $j^{\text{th}}$ sampled point along the given camera ray, $\phi_{i}$ denotes the $i^{\text{th}}$ FiLMed linear layer of the MLP followed by sinusoidal activation function (\ie Sine nonlinearity), and $\mathbf{x}^{(i)}_j$ is the input of $\phi_{i}$. $\boldsymbol{\gamma}^i$ and $\boldsymbol{\beta}^i$ are the affine transformation parameters of the FiLM layer~\cite{film}, which are the output of the mapping network conditioned on the latent vector $\mathbf{z} \in \mathbb{R}^{256}$.
Here, we implement the mapping network using an MLP of three hidden layers with 256 units each followed by leaky-ReLU activation with a slope of 0.2.
The FiLMed SIREN network is composed of eight FiLMed SIREN hidden layers of 256 units each.


\textbf{Normal prediction layer.}
We adopt the normal prediction layer that consists of MLP with two linear layers.
The first linear layer has 16 units followed by leaky-ReLU activation with a slope of 0.2. The second linear layer followed by Tanh activation and normalization to a unit vector maps the intermediate feature to the predicted normal $\mathbf{n} \in \mathbb{R}^3$.

\begin{figure*}[t]
    \centering
    \includegraphics[width=\linewidth]{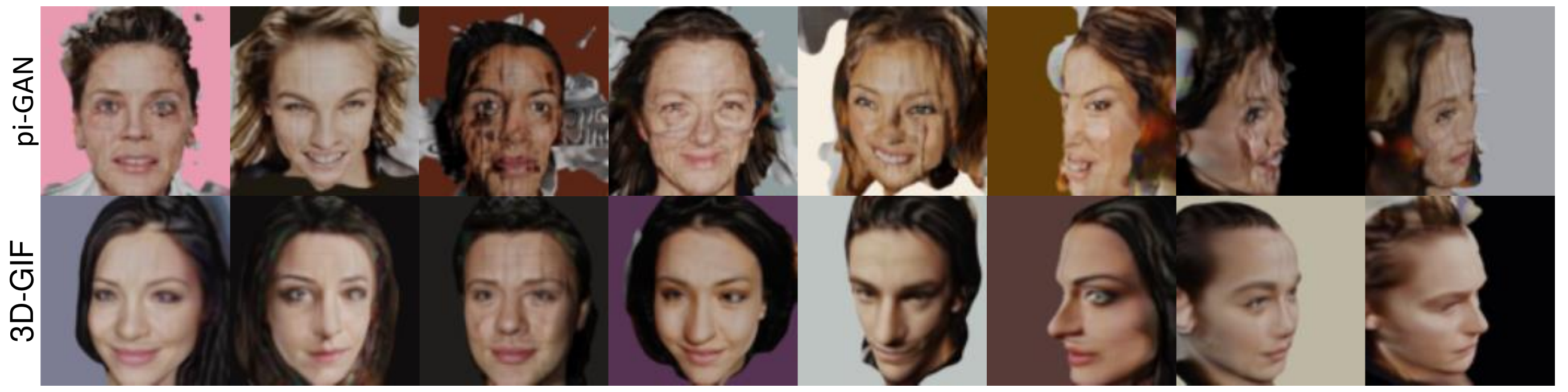}
    \vspace{-0.7cm}
    \caption{3D-rendered images from pi-GAN and our method. As illustrated, ours shows more realistic results than pi-GAN. Those images are utilized to measure the projection FID.}
    \label{fig:projection_fid}
\end{figure*}

\textbf{Discriminator.}
We leveraged CoordConv~\cite{coordconv} layers and residual connections~\cite{resnet} for the discriminator, referring to pi-GAN~\cite{pigan}. Unlike pi-GAN, we did not apply progressive discriminator.

\subsection{Training Details}
For the training, we utilized Adam optimizer with $\beta_1 = 0$ and $\beta_2 = 0.9$.
We set learning rates to $6\times10^{-5}$ and $2\times10^{-4}$ for the generator and the discriminator, respectively.
Also, we set $\lambda$ as 0.5 in Eq.~(\textcolor{red}{8}) and $\lambda_n$ as 20 in Eq.~(\textcolor{red}{11}) in our main paper. 
The number of iterations is set to 200,000 for the first stage and 20,000 for the second stage.
We conduct staged training where we first train Stage 1 and then jointly optimize Stage 1 and Stage 2. 
For stable training, hierarchical sampling~\cite{nerf} is adopted to our training procedure.
Also, we used Automatic Mixed Precision (AMP) in PyTorch for efficiency in computation and memory usage.
We utilize a pinhole perspective camera with a field of view (FOV) of $12^\circ$ for CelebA, CATs, and BFM.
Batch size and the number of samples along each ray are set to 56 and 12 for CelebA and BFM, and 28 and 24 for CATs.
For all datasets, the images are resized into $64\times64$ for the training.
Camera parameters and light condition parameters were adjusted according to each dataset, as described in Table~\ref{tab:trainingdetails}.


\begin{table}[h]
\resizebox{\linewidth}{!}{%
\begin{tabular}{c|ccc|cccc}
\toprule
\multirow{2}{*}{Dataset} & \multicolumn{3}{c|}{Camera param.} & \multicolumn{4}{c}{Light condition param.} \\
     & $dist_{cam}$ & $\sigma_v/\text{ range}_{v}$ & $\sigma_h/\text{ range}_{h}$ & $\mu_x$ & $\sigma_x$  & $\mu_y$ & $\sigma_y$  \\ \hline
CelebA  &  Gauss. & 0.15 & 0.3   & 0 & 0.27 & 0.39 & 0.07 \\
CATs    &  Uni. & (\text{-}0.5, 0.5)   & (\text{-}0.4, 0.4)   & 0 & 0.27 & 0.39 & 0.07 \\
BFM     &  Gauss. & 0.15   & 0.3   & 0 & 0.35  & 0 & 0.15    \\
\bottomrule
\end{tabular}%
}
\caption{Camera parameters and light condition parameters for each dataset. $dist_{cam}$, $v$, and $h$ denote camera pose distribution, pitch, and yaw. $x$ and $y$ denote the direction of light. Note that $\sigma_{v, h}$ in the camera parameters are for Gaussian distributions, and range$_{v, h}$ are for uniform distribution.}
\label{tab:trainingdetails}
\end{table}






\subsection{Albedo-Textured Mesh Extraction}
Since our model generates view-independent and light-disentangled representations, we can obtain high-quality albedo-textured meshes, which are easily applicable to the conventional computer graphics rendering pipelines.
To obtain the textured mesh, our model first predicts the volume density of each point in a $128\times128\times128$ voxel grid with the randomly-sampled latent vector $\mathbf{z}$ as a condition.
Here, we set a length of the voxel grid to include all the sampled points considering hyper-parameters (\eg, ray start, ray end, FOV, etc.) of each dataset.
Then, we extract a mesh (\ie, surface points and faces) from the volume density using a marching cube algorithm~\cite{marchingcube}.
Lastly, we texture the obtained mesh with the albedo predicted at each surface point by our model.
Unlike the previous NeRF-based methods~\cite{graf, pigan} that generate view-dependent and light-entangled colors, \ourmodel successfully generates meshes textured with albedo since our approach does not need view direction when predicting the albedo and separately controls the light condition.




\section{Experiment Details}
As quantitative evaluations, we compare the FID, depth error, and projection FID of our model and the baselines.
For each evaluation, we generated 2,000 images using each model.

\textbf{Depth error.}
To evaluate the quality of the generated geometry, we measure scale-invariant depth error (SIDE)~\cite{side} between the depth maps of the generated images and their ground truth.
As done in the previous approaches~\cite{depthnerf}, we extract depth maps of the generated images by volume rendering the distance between the given camera position and each sampled point.
In addition, to obtain the ground truth depth maps of the generated images, we leverage a depth estimation model~\cite{depth}.
We trained the depth estimation model with the training dataset of BFM.
Then, we estimated the ground truth depth map of each generated image using the pre-trained depth estimation model.

\textbf{Projection FID.}
As mentioned in Sec. \textcolor{red}{4.3} in our main paper, we measure projection FID of textured meshes extracted from ours and state-of-the-art model~\cite{pigan} as a quantitative comparison.
Here, projection FID indicates the FID score of projected images from the textured meshes.
We first extract surface meshes using a marching cube algorithm~\cite{marchingcube}.
Since the estimated surface changes according to the sigma threshold of the algorithm, we extract meshes at different thresholds from 10 to 200 with an interval of 10. 
Then, we texture the meshes with the generated colors.
Since the baseline model requires view direction to predict colors, we provide the model with the frontal view direction as a condition.
To project images from the textured mesh, we utilize Blender~\cite{blender} with sampled viewpoints and light conditions following our training setting.
Moreover, we added random solid colors as backgrounds to the projected images since they lack valid backgrounds, unlike the real images.
Table~\textcolor{red}{3} in our main paper presents projection FID scores according to different sigma thresholds and the best score of each model. 
Fig.~\ref{fig:projection_fid} presents the examples of projected images of ours and the baseline at the sigma threshold of the best score.
\vspace{-0.3cm}
\section{Additional Experiments}
\vspace{-0.2cm}

\begin{figure}[h!]
    \centering
    \includegraphics[width=\linewidth]{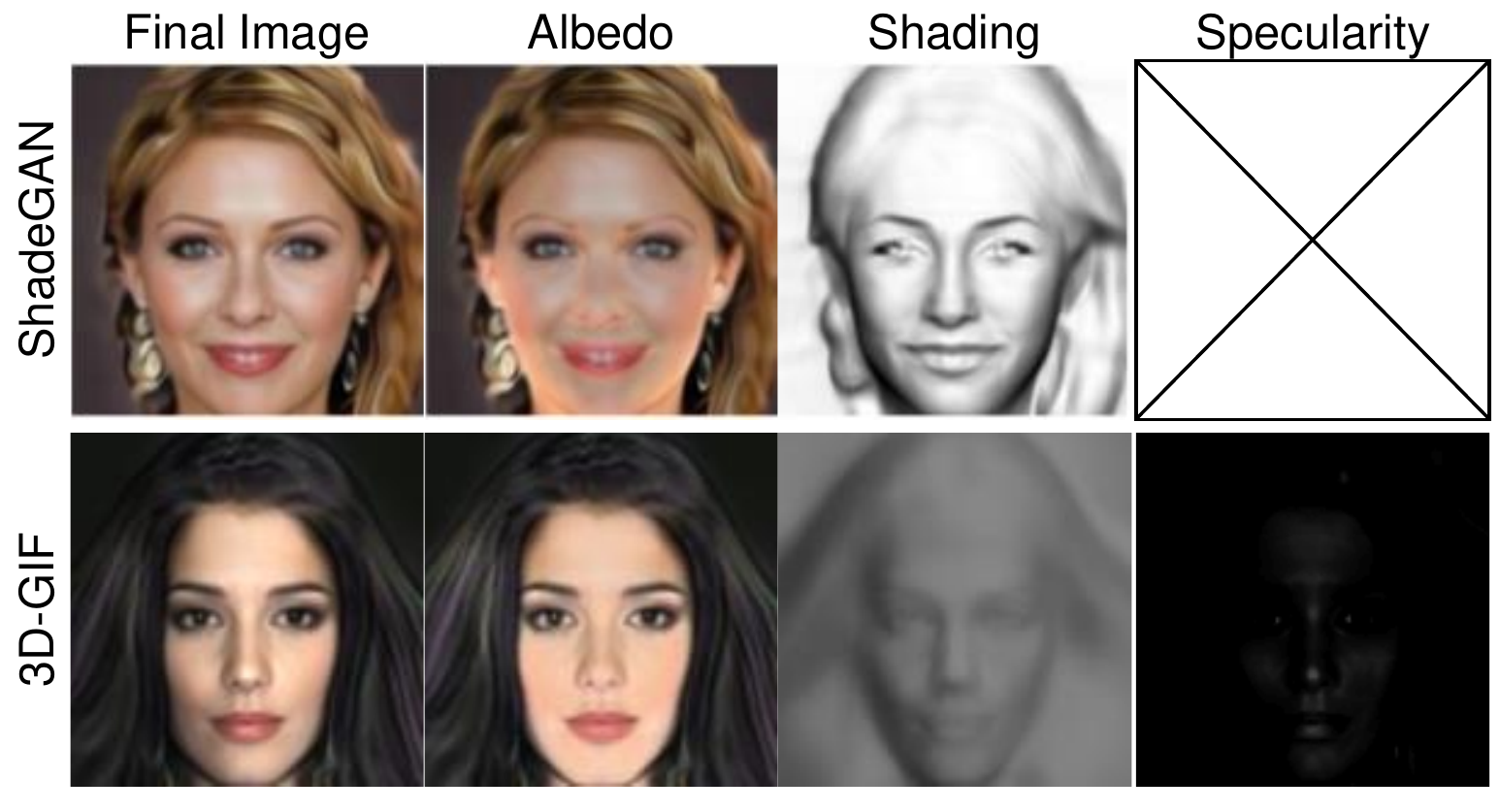}
    \vspace{-0.7cm}
    \caption{Visualization of factorized representations from ShadeGAN and our method. Generated albedo from ShadeGAN contains specularity while ours does not.}
    \label{fig:shadegan_factor}
\end{figure}

\begin{figure*}[h!]
    \centering
    \includegraphics[width=\linewidth]{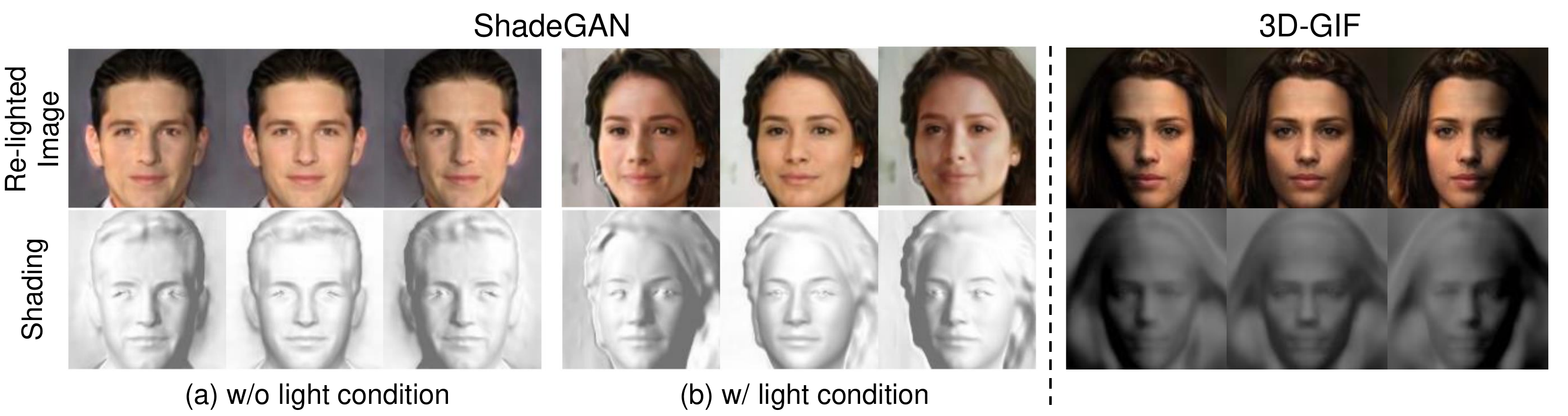}
    \vspace{-0.7cm}
    \caption{Re-lighted images from ShadeGAN and our method. As shown, our model reflects the light changes more effectively than ShadeGAN. The ShadeGAN presents two settings, whether light condition is concatenated to intermediate feature or not. (a) images are generated without light concatenation, and (b) images are generated with light concatenation.}
    \label{fig:shadegan_relight}
\end{figure*}

\begin{figure*}[h!]
    \centering
    \includegraphics[width=\linewidth]{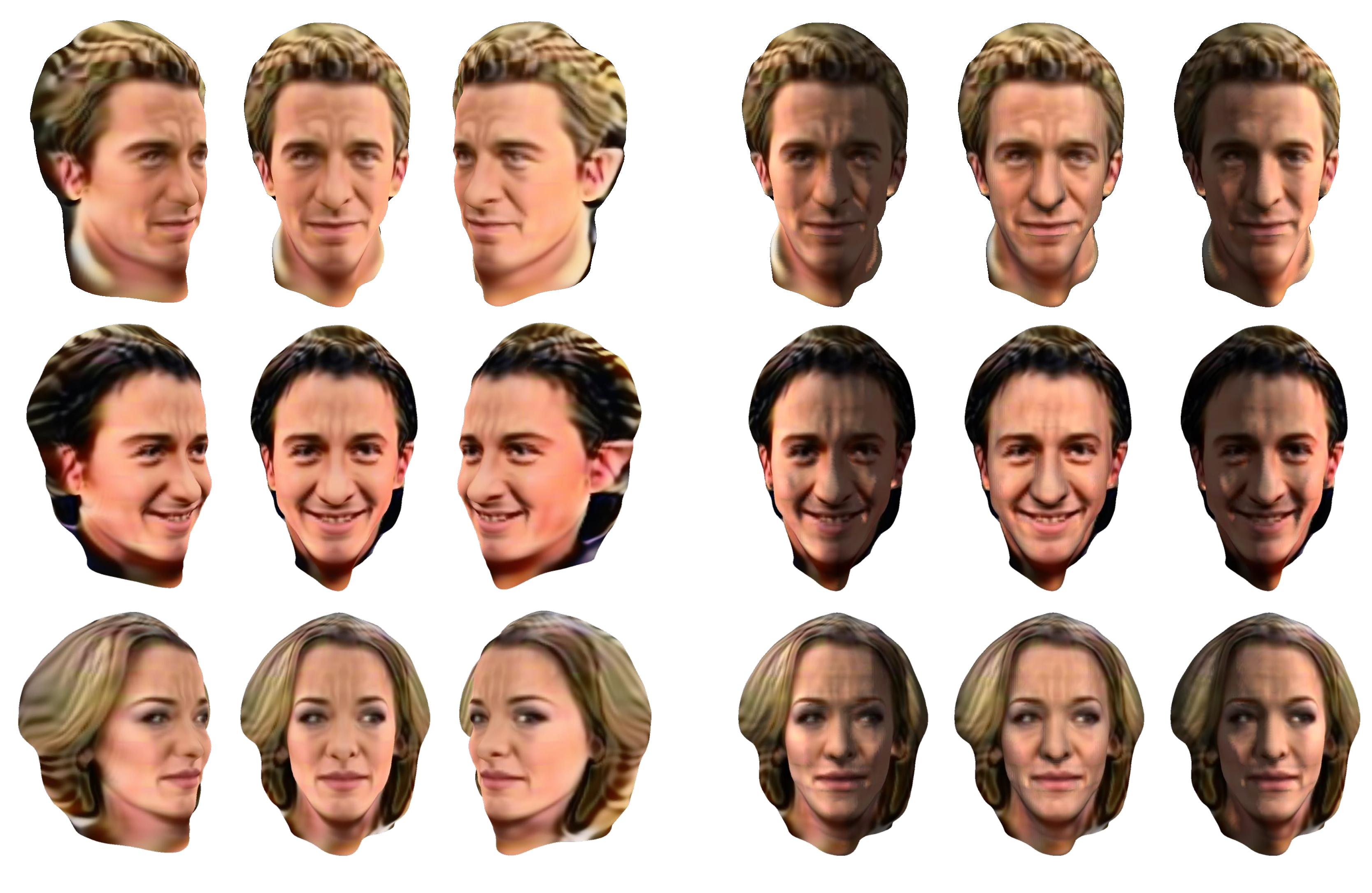}
    \vspace{-0.7cm}
    \caption{Our generated albedo-textured meshes and their re-lighted results on CelebA dataset.}
    \label{fig:mesh_celeba}
\end{figure*}
\begin{figure*}[h!]
    \centering
    \includegraphics[width=\linewidth]{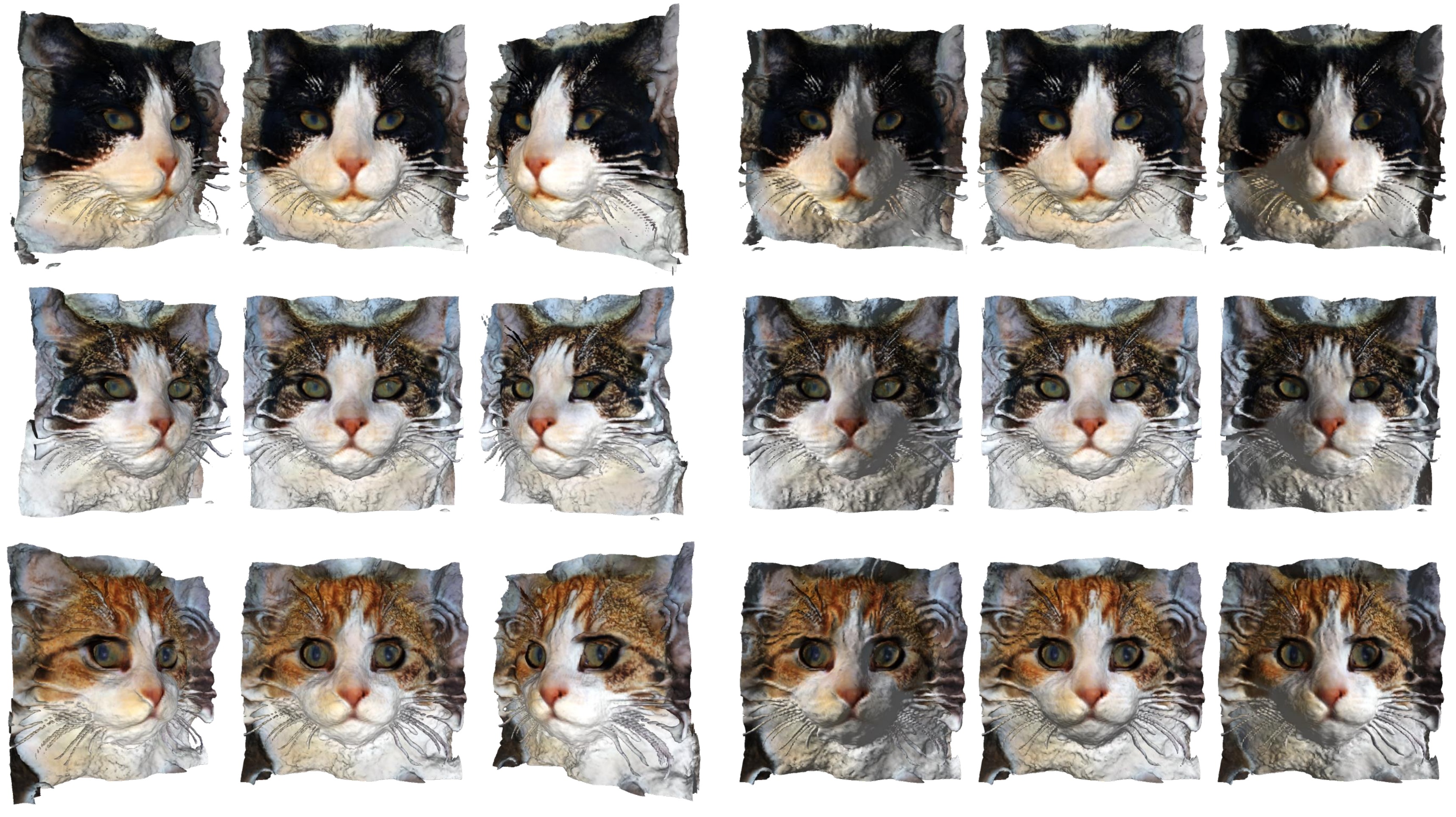}
    \vspace{-0.7cm}
    \caption{Our generated albedo-textured meshes and their re-lighted results on CATs dataset.}
    \label{fig:mesh_cat}
\end{figure*}

We report additional qualitative results using CelebA and CATs datasets.
Fig.~\ref{fig:mesh_celeba} and Fig.~\ref{fig:mesh_cat} present our generated albedo-textured meshes and their re-lighted results.
As shown, our model generates realistic albedo-textured meshes applicable to the conventional rendering pipeline.

\begin{figure*}[h!]
    \centering
    \includegraphics[width=\linewidth]{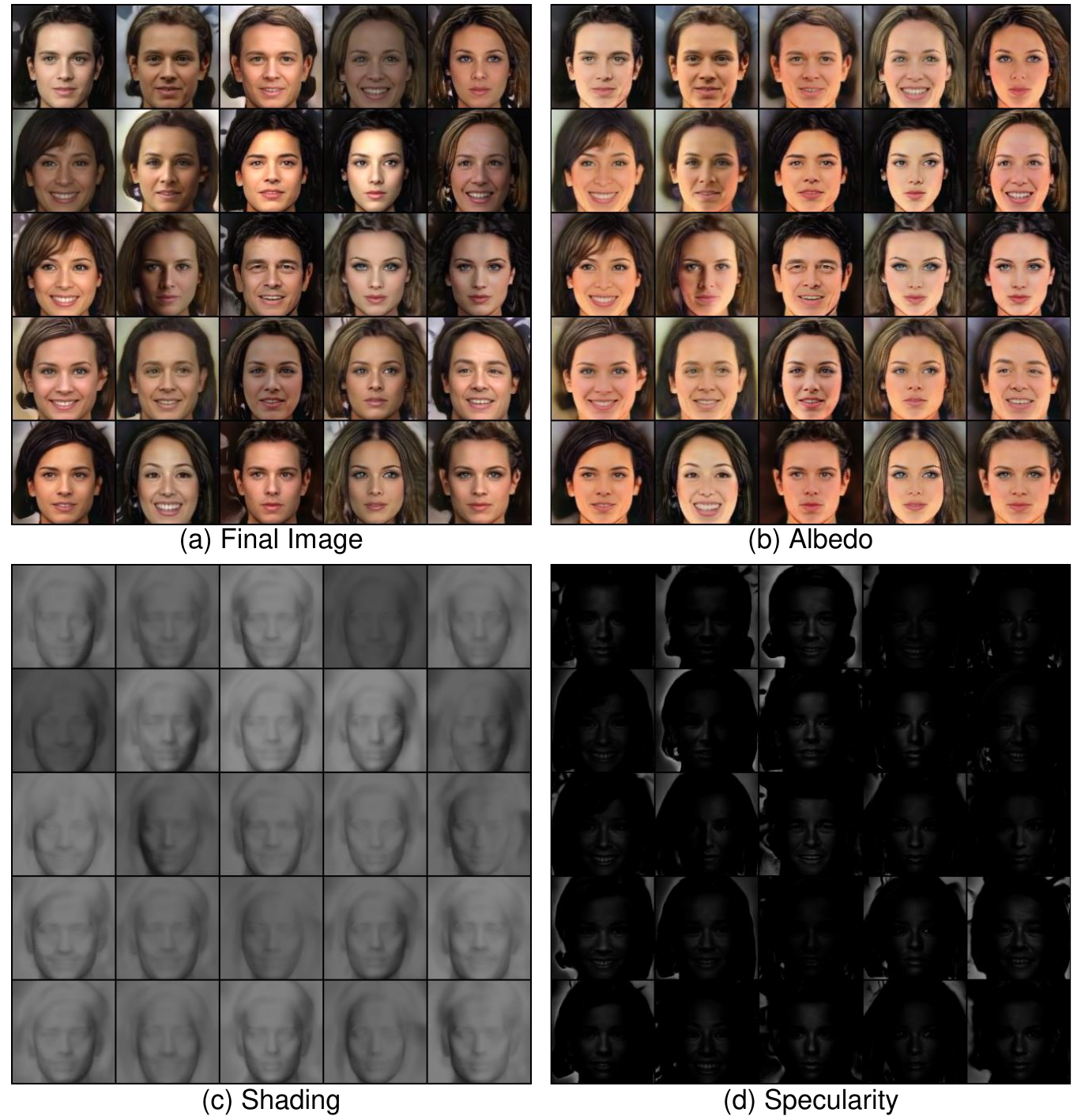}
    \vspace{-0.7cm}
    \caption{Random samples of CelebA. (a), (b), (c), and (d) indicate final images, albedo, shading, and specularity, respectively.}
    \label{fig:random_celeba}
\end{figure*}

\begin{figure*}[h!]
    \centering
    \includegraphics[width=\linewidth]{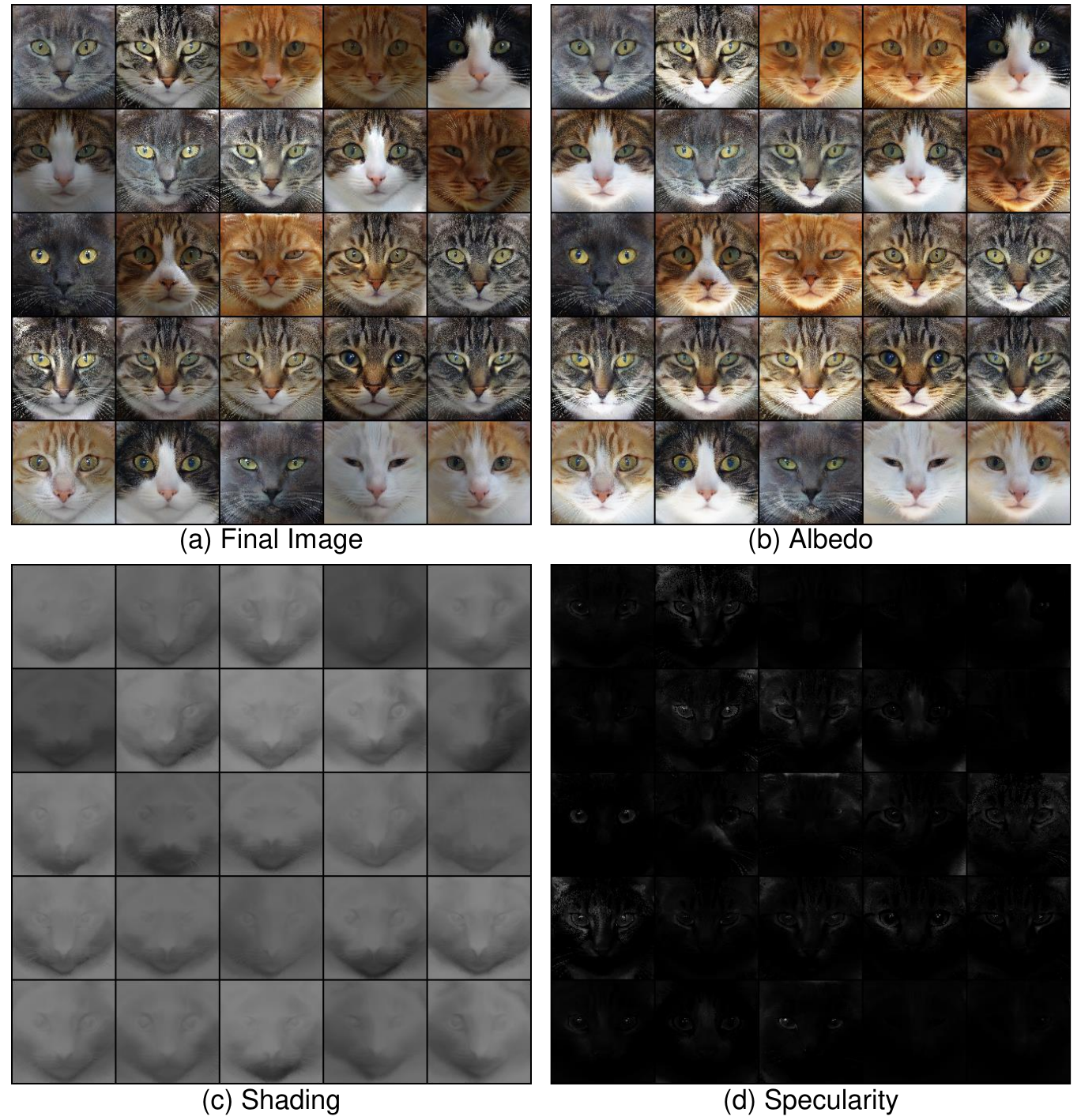}
    \vspace{-0.7cm}
    \caption{Random samples of CATs. (a), (b), (c), and (d) indicate final images, albedo, shading, and specularity, respectively.}
    \label{fig:random_cats}
\end{figure*}

To further illustrate our robustness in generating high-quality factorized representations, we show randomly sampled results in Fig.~\ref{fig:random_celeba} and Fig.~\ref{fig:random_cats}. 
Note that those results are obtained by varying a random seed from 0 to 25.




\section{Comparison with ShadeGAN}
ShadeGAN~\cite{shadegan}, which is concurrent work presented in arXiv recently, proposes a framework to improve the quality of geometry from the previous NeRF-based generative methods by generating light-disentangled albedo under the Lambertian assumption.
While albedo is independent of view direction by its definition, ShadeGAN predicts view-dependent albedo conditioned on a certain view direction.
This view dependency enables the model to express specularity even under the Lambertian shading assumption where the specular reflection is not involved.
However, as also described in Sec.~\textcolor{red}{3.2} of our main paper, the view dependency of the generated volume representation hampers its applicability in various 3D applications.
This is because they still need redundant rendering for every desired viewpoint, and their albedo contains specularity.
Although ShadeGAN improved the geometry quality, as shown in Fig.~\ref{fig:shadegan_factor} and Fig.~\ref{fig:shadegan_relight}, their generated albedo image still contains specularity, and their re-lighted images cannot fully reflect the given light conditions.
Note that we brought the result images of ShadeGAN from their paper to compare with ours since their official code is not publicly available yet.

\end{document}